\documentclass{article}

\usepackage{microtype}
\usepackage{graphicx}
\usepackage{subfigure}
\usepackage{booktabs} 

\usepackage{hyperref}

\usepackage[accepted]{icml2025}

\usepackage{amsmath}
\usepackage{amssymb}
\usepackage{mathtools}
\usepackage{amsthm}

\usepackage[capitalize,noabbrev]{cleveref}

\theoremstyle{plain}

\theoremstyle{definition}

\theoremstyle{remark}

\usepackage[textsize=tiny]{todonotes}

\icmltitlerunning{Modeling User Behavior from Adaptive Surveys with Supplemental Context }

\begin{document}

\twocolumn[
\icmltitle{Modeling User Behavior from Adaptive Surveys with Supplemental Context}




\begin{icmlauthorlist}
\icmlauthor{Aman Shukla}{comp}
\icmlauthor{Daniel Patrick Scantlebury}{comp}
\icmlauthor{Rishabh Kumar}{comp}
\end{icmlauthorlist}

\icmlaffiliation{comp}{Resonate Networks, Inc., Virginia, USA}

\icmlcorrespondingauthor{Aman Shukla}{aman.shukla@resonate.com}

\icmlkeywords{Hybrid User Modeling, Tabular Structured Data}

\vskip 0.3in
]



\printAffiliationsAndNotice{}  

\begin{abstract}
Modeling user behavior is critical across many industries where understanding preferences, intent, or decisions informs personalization, targeting, and strategic outcomes. Surveys have long served as a classical mechanism for collecting such behavioral data due to their interpretability, structure, and ease of deployment. However, surveys alone are inherently limited by user fatigue, incomplete responses, and practical constraints on their length making them insufficient for capturing user behavior. In this work, we present LANTERN (Late-Attentive Network for Enriched Response Modeling), a modular architecture for modeling user behavior by fusing adaptive survey responses with supplemental contextual signals. We demonstrate the architectural value of maintaining survey primacy through selective gating, residual connections and late fusion via cross-attention, treating survey data as the primary signal while incorporating external modalities only when relevant. LANTERN outperforms strong survey-only baselines in multi-label prediction of survey responses. We further investigate threshold sensitivity and the benefits of selective modality reliance through ablation and rare/frequent attribute analysis. LANTERN's modularity supports scalable integration of new encoders and evolving datasets. This work provides a practical and extensible blueprint for behavior modeling in survey-centric applications.
\end{abstract}

\section{Introduction}
Understanding user behavior is a fundamental challenge in applied machine learning. In domains like marketing, healthcare, and public policy, this often takes the form of measuring preferences, decisions, and experiences over time. Surveys have historically played a key role here due to their structured format, interpretability, and wide adoption. Whether embedded in feedback loops, onboarding flows, or customer lifecycle touchpoints, surveys provide domain-specific and longitudinal insight into user traits. However, the practical limitations of surveys are well known. Users cannot be expected to engage with long or frequent questionnaires. The result is often a tradeoff between breadth and depth: surveys either ask fewer questions to maximize response rates or risk high dropout and noise. As a result, important aspects of user behavior may remain unobserved. Crucially, survey responses are bounded by who chooses to answer, creating a structured but inherently sparse view of the population.

To counter this, many systems now rely on hybrid modeling strategies \cite{zhang2023cooccurrencemultimodalsessionbasedrecommendation}, \cite{Wei_2023}, supplementing primary signals with contextual or observational data. In contrast to surveys, supplementary signals, such as demographic traits, engagement metrics, or transactional logs are collected passively in the wild. These signals are not constrained by voluntary response and thus exist in much higher volume. However, the massive volume also introduces significant noise and potential bias making them volatile inputs in hybrid modeling. The challenge is to combine these supplementary features with survey responses in a way that respects the structure and semantics of the survey, avoids feature bloat, and remains interpretable and modular.

\begin{figure}[ht]
\vskip 0.2in
\begin{center}
\centerline{\includegraphics[width=\columnwidth]{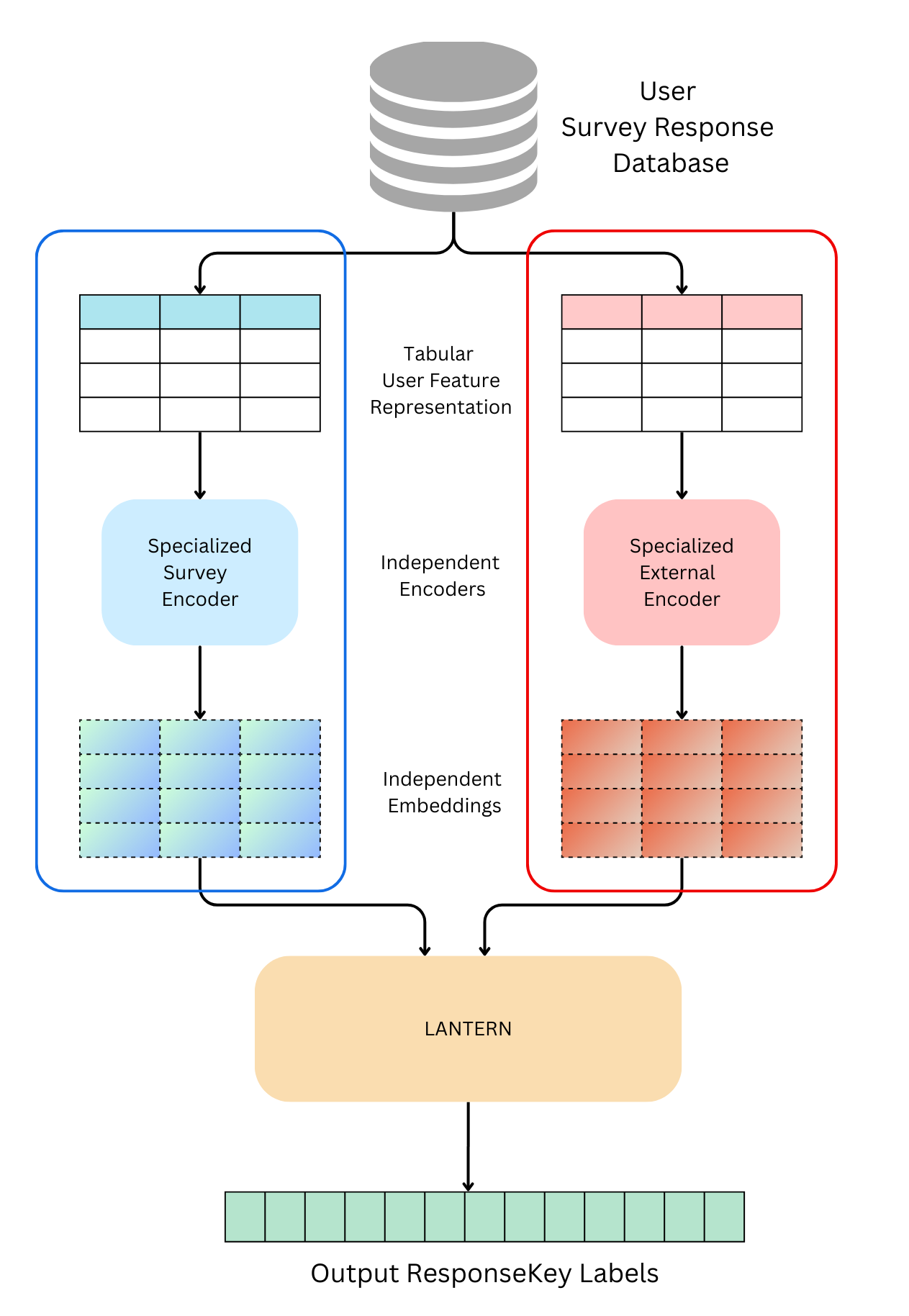}}
\caption{Ecosystem Overview. (Left) Pipeline for generating signal $x_s$ from survey. (Right) Pipeline to generate signal $x_e$ from external data. Both pipelines sample information from a database of survey respondents that includes different modalities including user behavior.}
\label{hm-overview}
\end{center}
\vskip -0.2in
\end{figure}

Our model LANTERN, is anchored on a large, user-facing survey that serves as a critical source of ground truth. This survey captures a broad spectrum of behavioral, demographic, and psychographic attributes, including but not limited to age, gender, ethnicity, brand preferences, digital media habits, lifestyle interests, and purchasing intentions. By sourcing these responses directly from users at the time of survey deployment, we obtain high fidelity labels that reflect self-reported truths, distinguishing this signal from passively observed or inferred behavioral data. The richness and scope of this survey design empower us to build nuanced behavioral representations and ensure our learning framework is guided by concrete signals with deep relevance to user understanding, personalization, and market insights. This anchoring mechanism also positions the survey as an indispensable foundation upon which supplemental behavioral signals can be integrated to expand coverage and depth. LANTERN is a scalable neural architecture designed to keep survey signals central while selectively integrating external context. The model applies late fusion via cross-attention, using survey embeddings to query contextual embeddings and learn only behaviorally relevant interactions from supplementary signals. The gating and residual mechanisms preserve interpretability and alignment with the survey semantics. Furthermore, both survey and supplemental data are stored in structured tabular formats within data warehouses in the industry. We provide a blueprint, illustrated in Figure \ref{hm-overview}, in which we process these tables through encoder modules that produce embeddings for each user. These embeddings are then passed to LANTERN. Importantly, the system is built for deployment: it is modular, efficient ($\sim$50M parameters), and ready for deployment.

\section{Related Work}

Recent advancements in deep learning for user modeling have demonstrated the utility of structured behavioral signals. For example, Sukhbaatar et al. \cite{Sukhbaatar_Usagawa_Choimaa_2019} apply a multi-layer perceptron (MLP) to prediction factors derived from student's online learning activities to identify those at risk of course failure. Gao et al. \cite{gao} introduce a cascading assumption among behavior types and integrate Neural Collaborative Filtering \cite{ncf} within a multi-task learning framework. Despite their effectiveness, these approaches generally assume behavioral signals are self-sufficient, often overlooking supplementary contextual cues.
Two-step user modeling architectures such as DMT \cite{dmt} and ZEUS \cite{zeus} learn cross-type user representations by concatenating intra-type sequential embeddings, while GHCF conducts multi-behavior prediction via behavior-wise fusion of intra-type collaborative representations \cite{ghcf}. However, these methods fall short in modeling interdependent context across modalities.

Attention mechanisms have emerged as powerful tools for multimodal modeling. Tsai et al. \cite{tsai-etal-2019-multimodal} introduce a transformer-based architecture that performs cross-modal attention without requiring temporal alignment, enabling effective fusion of language, visual, and acoustic signals for sentiment and emotion tasks. ViLBERT \cite{vibert} extends the BERT \cite{devlin2019bertpretrainingdeepbidirectional} model with a two-stream architecture, applying co-attentional transformers to integrate visual and textual inputs, and achieves state-of-the-art results across vision and language benchmarks.
Gating mechanisms have also been shown to be effective in handling task and modality heterogeneity. The Multi-gate Mixture-of-Experts (MMoE) framework \cite{mmoe} supports task-specific routing of shared experts, yielding strong gains in recommendation and advertising contexts. Likewise, the Gated Multimodal Unit (GMU) \cite{gmu} introduces a learnable gating structure for fusing multimodal inputs, demonstrating improvements on genre classification using plot and poster data.

Our work is inspired by these advances in fusion strategies, gated learning, and cross-modal attention. However, unlike prior work, which often targets high-dimensional or richly aligned modalities, we focus on the underexplored setting of leveraging these techniques on heterogeneous tabular user behavior data in advertising. In doing so, we bridge the gap between state-of-the-art user modeling techniques and practical, industry-scale tabular data environments.

\section{Methodology}
\subsection{Problem Formulation}
Let $x_s$ denote adaptive survey responses and $x_e$ represent external contextual features for $N$ users. Each user completes an adaptive survey composed of multiple-choice questions of three types - binary choice, single choice, and multiple choice. We model this at the response key level: each possible answer option is treated as an independent binary label. For binary choice, we model only one key (e.g. “Yes”), which is set to 1 if selected, 0 otherwise. Let $y \in \{0,1\}^{N \times d}$ be the multi-label target matrix, where $d$ is the number of distinct response keys.

\begin{figure}[ht]
\vskip 0.2in
\begin{center}
\centerline{\includegraphics[width=\columnwidth]{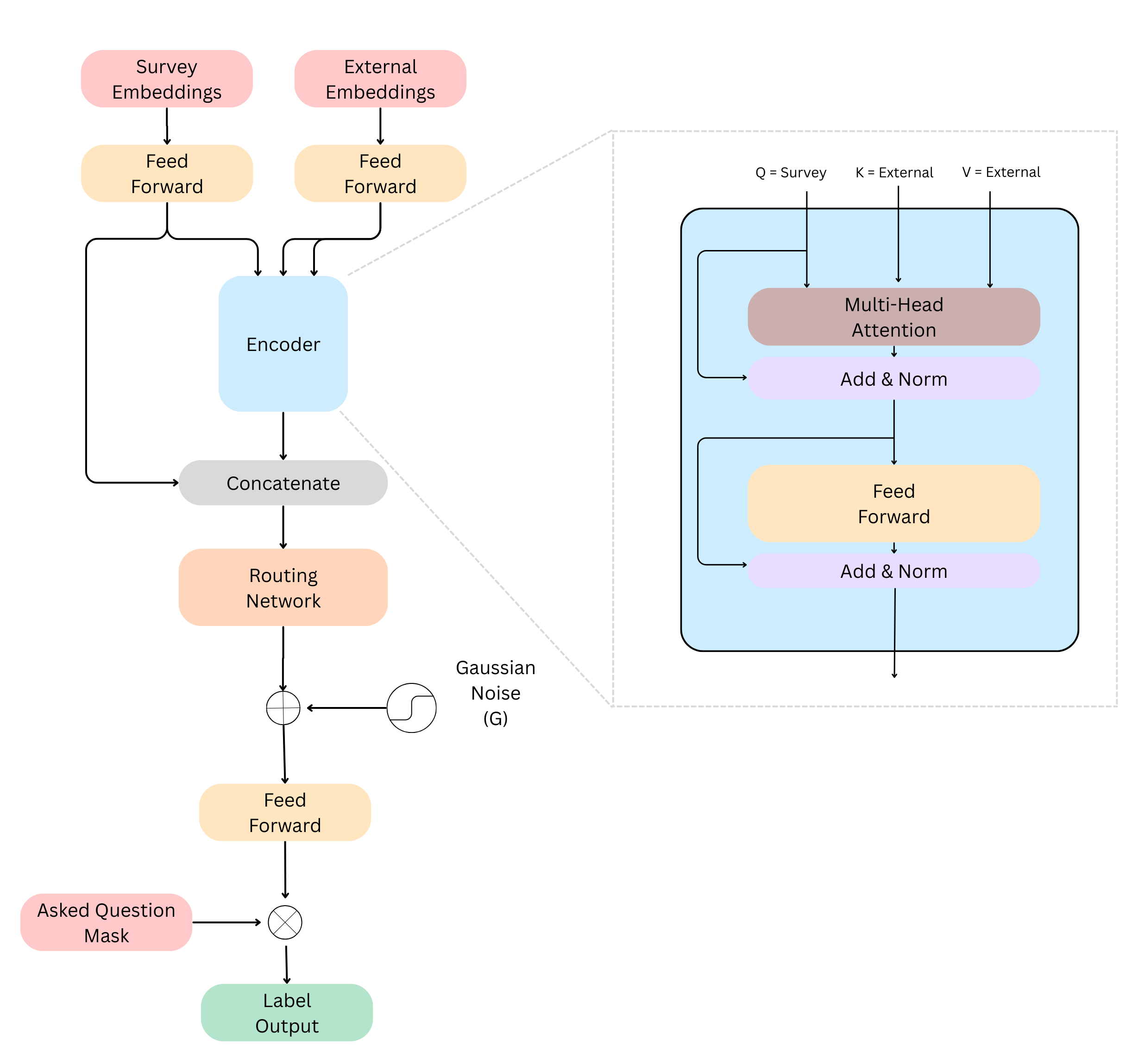}}
\caption{LANTERN with cross-attention and late fusion.}
\label{hm-architecture}
\end{center}
\vskip -0.2in
\end{figure}

\subsection{Architecture}
\subsubsection{Encoder Layers}
Inputs $x_s \in \mathbf{R}^{N \times F_s}$ and $x_e \in \mathbf{R}^{N \times F_e}$ are encoded separately:
\begin{align*}
    {h_s}^{N \times D} = f_s(x_s) \\
   {h_e}^{N \times D} = f_e(x_e)
\end{align*}
where $F_s, F_e$ are features in the survey and external datasets. $f_s,f_e$ are learned encoders for the survey and external data respectively. $h_s, h_e$ are corresponding embeddings of size $N \times D$. 

\subsubsection{Cross Attention Layer}
The encoded survey representation $h_s$ is then passed through a Transformer \cite{transformer} block, which takes external embeddings $h_e$ as both key and value as shown in Figure \ref{hm-architecture}. The result is a contextualized embedding $h_{t}$ , which reflects relevant external features attended to by the survey input.
\begin{align*}
    h_t = \textbf{Encoder}(Q=h_s, K=h_e, V=h_e)
\end{align*}

\subsubsection{Gated Residual Fusion}
We compute the fused embeddings as 
\begin{align*}
    h_{fused} = h_s + g \odot (h_t - h_s)
\end{align*}
where $g \in (0,1)$ is a learned gate and $\odot$ denotes element wise multiplication. We add Gaussian noise ($G$) as a regularizer to the signal to prevent overfitting.

\subsubsection{Output Head \& Loss}
The noisy fused embedding is passed through a feed-forward layer with sigmoid activation, yielding per-key probabilities $\hat{y} \in [0,1]^{N \times d_s}$. A mask $m \in \{-1,0,1\}^{N \times d_s}$ is applied to extract the ground truth. Binary Cross Entropy is used to calculate loss against the extracted labels.
\begin{align*}
    \textit{\emph{L}} = \textbf{BinaryCrossentropy}(m \odot \hat{y})
\end{align*}

\subsection{Implementation and Scaling}
LANTERN is implemented in TensorFlow and has $\sim$50M parameters. Each encoder is modular and versioned independently. Cross-attention uses 8 heads, and residual fusion includes dropout and LayerNorm. Gaussian noise acts as a regularizer.

Components of LANTERN are deployed in production, where it supports real-time inference. Because encoders are decoupled, they can be improved or retrained independently. Late fusion ensures graceful degradation when external data is missing or delayed. The modular architecture means that new modalities can be added via plug-in encoders.

\section{Experimental Setup}

\subsubsection{Dataset}
Our dataset comprises $\sim$35,000 anonymized users from a production-grade survey system. Users completed adaptive surveys spanning binary, single, and multiple-choice questions. External dataset is acquired from distributors in the field and aligned at the user level.

\subsubsection{Evaluation Framework}
We evaluate our experiment in four different ways.
\begin{itemize}
    \item[(i)] \textbf{Core Predictive Metrics}: Since we setup the problem as a supervised classification, we monitor metrics like Precision, Recall and F1-Score. Accuracy and ROC-AUC are dropped because of the sparse nature of the problem. 
    \item[(ii)] \textbf{Ablation Study}: Comparing LANTERN with survey-only and external only models
    \item[(iii)] \textbf{Frequent vs Rare Attributes}: To gain deeper understanding of LANTERN's behavior, we examine it's performance on two segments - frequent (highly observed responses) and rare (sparsely observed responses).
    \item [(iv)] \textbf{Thresholding Analysis}: We include a thresholding analysis to examine how precision, recall, and F1-score vary with the decision threshold. The choice of threshold can significantly affect the metric values, especially when the label distributions are imbalanced. This analysis serves as a diagnostic to assess calibration and stability of model outputs and allows practitioners to make more informed decisions about recall-oriented vs. precision-oriented configurations. 
\end{itemize}


\begin{table}[t]
\caption{Scores for different setups in the ablation study.}
\label{ablation-table}
\vskip 0.15in
\begin{center}
\begin{small}
\begin{sc}
\begin{tabular}{lccccr}
\toprule
Setup & Precision  & Recall & F1-Score  \\
\midrule
Survey-only    & 0.7976 & 0.6794 & 0.7338 \\
External-only & 0.7537 & 0.4264 & 0.5447\\
LANTERN    & \textbf{0.8263} & \textbf{0.7296} & \textbf{0.7750} \\
\bottomrule
\end{tabular}
\end{sc}
\end{small}
\end{center}
\vskip -0.1in
\end{table}

\begin{table*}[t]
\caption{Rare vs Frequent Response Analysis. 1000 least responded keys are bucketed as $\it{rare}$ and 1000 most responded keys are bucketed as $\it{frequent}$}
\label{freq-table}
\vskip 0.15in
\begin{center}
\begin{small}
\begin{sc}
\begin{tabular}{lcccccc}
\toprule
Setup & \multicolumn{3}{c}{Rare Attributes} & \multicolumn{3}{c}{Frequent Attributes} \\
\cmidrule(lr){2-4} \cmidrule(lr){5-7}
 & Precision & Recall & F1 & Precision & Recall & F1 \\
\midrule
Survey   & 0.8755 & 0.8161 & 0.8448 & 0.7865 & 0.6165 & 0.6912 \\
External & 0.7776 & 0.5932 & 0.6730 & 0.7931 & 0.6029 & 0.6850 \\
LANTERN & 0.8751 & \textbf{0.8404} & \textbf{0.8575} & 0.7901 & \textbf{0.6484} & \textbf{0.7123} \\
\bottomrule
\end{tabular}
\end{sc}
\end{small}
\end{center}
\vskip -0.1in
\end{table*}

\section{Results \& Discussion}

The results shown below were performed on a sample set of 35,000 users.
\subsection{Core Predictive Metrics}
The metrics for LANTERN's performance is shown in the last row of Table \ref{ablation-table}
\subsection{Ablation Results}

LANTERN consistently outperforms both the survey-only and external-only baselines (shown in Table \ref{ablation-table}). Notably, while the survey-only model is relatively strong (F1 $\approx$ 0.73) due to the informative structure of adaptive questionnaires and poor performance with the external information as the only source; LANTERN combines useful information from the external data and enriches the original information significantly improving recall by 5 points.


\subsection{Rare vs Frequent Attribute Analysis}
This segment reflects real-world survey dynamics, where some questions or options are shown conditionally and others are shown universally. Evaluating by frequency helps validate whether LANTERN effectively adapts to both highly specific and broadly distributed behavioral signals. 
The results are tabulated in Table \ref{freq-table}. On rare attributes, the survey-only model performs well (F1 $\approx$ 0.84) due to the precise targeting and high signal-to-noise ratio of conditionally served questions. LANTERN improves on this with higher recall, yielding an F1-score of 0.8575. This suggests that the external context is leveraged judiciously to recover rare selections that may not be explicitly queried.
On frequent attributes, both survey-only and external-only models perform worse due to increased noise and ambiguity. LANTERN again shows improvement (F1 = 0.7123), driven primarily by gains in recall. This indicates that late fusion helps synthesize overlapping signals in noisy, high-frequency settings.

\subsection{Thresholding Analysis}
\begin{figure}[ht]
\vskip 0.2in
\begin{center}
\centerline{\includegraphics[width=\columnwidth]{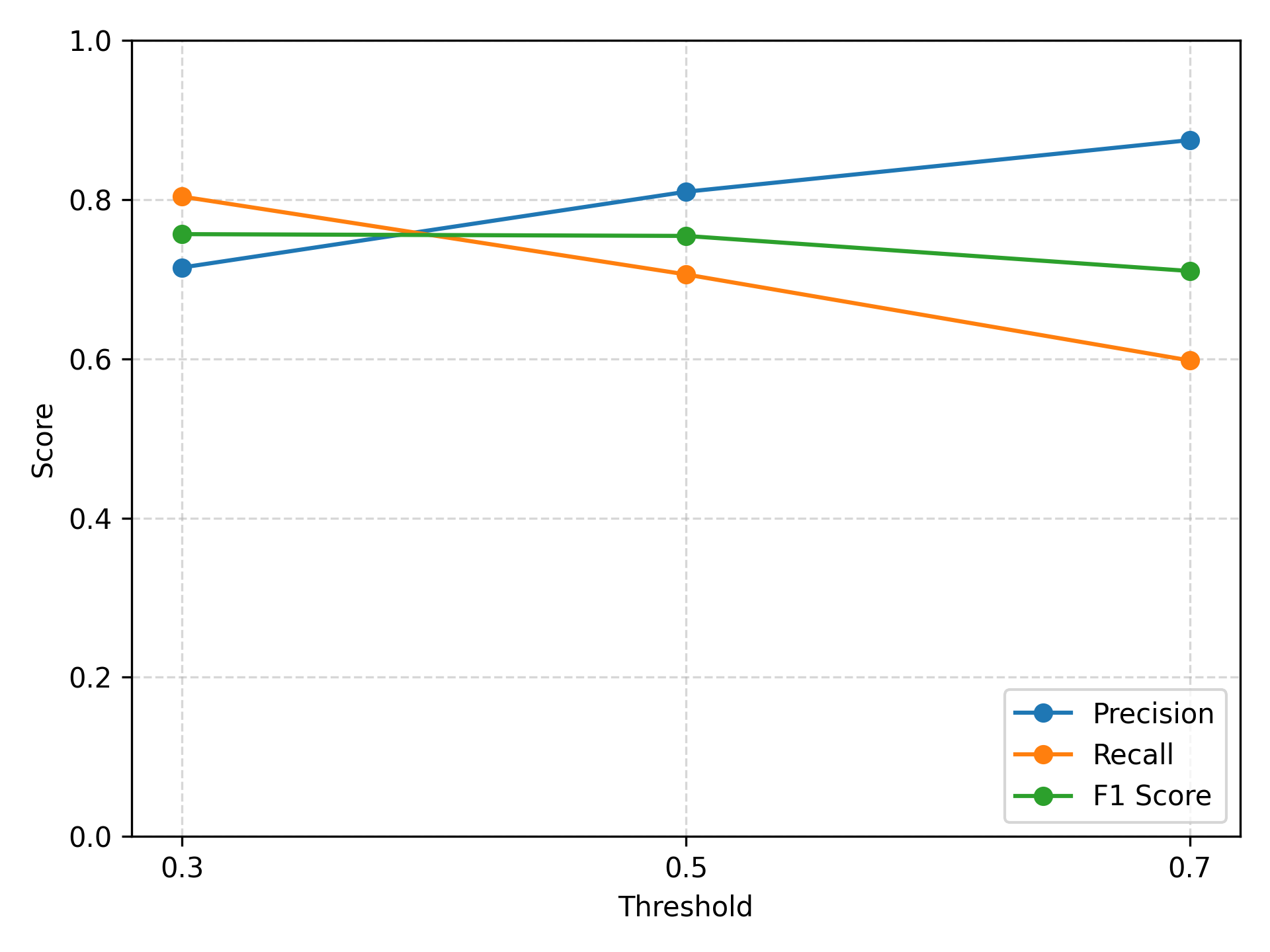}}
\caption{Threshold sensitivity of LANTERN evaluated at decision thresholds- 0.3, 0.5, 0.7. }
\label{threshold}
\end{center}
\vskip -0.2in
\end{figure}

Figure \ref{threshold} illustrates how LANTERN’s performance metrics vary across thresholds. As the threshold increases, precision improves steadily, while recall drops. The F1-score remains comparatively stable ($\sim$0.75). These trends reflect the typical trade-offs in multi-label prediction: lower thresholds recover more positives (higher recall) but reduce confidence (lower precision), while higher thresholds reduce false positives at the cost of under-prediction. The relatively flat F1-score curve suggests that LANTERN’s output probabilities are reasonably well calibrated, but this analysis also highlights the potential value of incorporating a threshold tuning procedure to better tune response keys.

\section{Conclusion and Future Work}

LANTERN is a scalable, interpretable model for behavior prediction from adaptive survey data. It strikes a practical balance between data richness and architectural simplicity by leveraging late-fusion cross-attention. We believe this approach provides a solid foundation for future research on hybrid behavior modeling in resource-constrained, survey-centric environments. 

Looking ahead, one of the most pressing next steps is the development of a threshold tuning framework tailored to the characteristics of this task. Because each response key exhibits a different label distribution; we plan to implement response-key-specific threshold optimization that adjusts decision boundaries using observed distributions and validation signal, with the goal of improving both calibration and downstream utility in production systems. Additional future directions include temporal generalization across survey cycles, multi-task extensions to model grouped behaviors jointly, and plug-and-play adaptation to emerging data modalities such as interaction logs.

\section*{Acknowledgements}
We thank Dan Ensign and Robert Reaney for their invaluable insights early on that helped shape the core design of LANTERN. The authors also thank Mustafa Kanchwala and Pooja Giri for their continued efforts to deploy and maintain the system infrastructure. Their work in optimizing LANTERN's components for compute resource availability and production stability was instrumental in bringing the model to scale.

\bibliography{example_paper}
\bibliographystyle{icml2025}

\newpage
\appendix
\onecolumn 
\section{Implementation Details}
LANTERN is implemented in TensorFlow and trained using the Keras functional API. The model consists of two encoders, one for survey features, and one for external context followed by a transformer based fusion block, a gated residual connection, and a prediction head with dynamic masking.

\begin{itemize}
    \item Model Size: $\sim$50 million parameters
    \item Survey Encoder: Produces embeddings of size 512 for each user which are projected to 2048 units using a feed-forward network.
    \item External Encoder: Symmetric to the survey encoder, produces embeddings of size 512 for each user which are projected to 2048 units using a feed-forward network.
    \item Encoder Block: We reshape the projected embeddings to (BATCH, 64, 32) for efficient Multi-head attention. We use two feed-forward layers in the block; first to project the high dimensional embeddings to 4096 dimension hidden layer and then back to 512 for repetition. We use 3 layers of transformer blocks based on preliminary experiments to gain maximum benefits of attention.
    \item Gating: Learned via feed-forward network with sigmoid activation.
    \item Output Label and Mask: Dense layer with $d_s$ units to match the unique survey response keys. A mask $m \in \{-1,0,1\}$ depending on whether the response key is non-favorable, wasn't asked, or favorable respectively is applied. This masking is done to calibrate the loss function and inform learning from ground truth (favorable/non-favorable) only; and discard contributions if the response key wasn't asked to a user.
    
\end{itemize}

The training configuration of LANTERN is provided below, 
\begin{itemize}
    \item Optimizer : Adam with default settings of $\beta_1 = 0.9$, $\beta_2 = 0.99$, and learning rate of $1e-03$. 
    \item Loss: Binary Cross Entropy
    \item Batch Size: 256
    \item Epochs: Trained to 30 epochs with $\emph{training\_steps\_per\_epoch}=1000$ and $\emph{validation\_steps\_per\_epoch}=100$. The data was loaded with \emph{repeat} enabled to prevent data exhaustion; \emph{shuffle} was turned on to discourage any ordering.
    \item Hardware: LANTERN was trained for 4 hours on a single A10G instance GPU using AWS Cloud Infrastructure.
\end{itemize}

\section{Gating Behavior Analysis}
\begin{figure}[ht]
\vskip 0.2in
\begin{center}
\centerline{\includegraphics[width=\columnwidth]{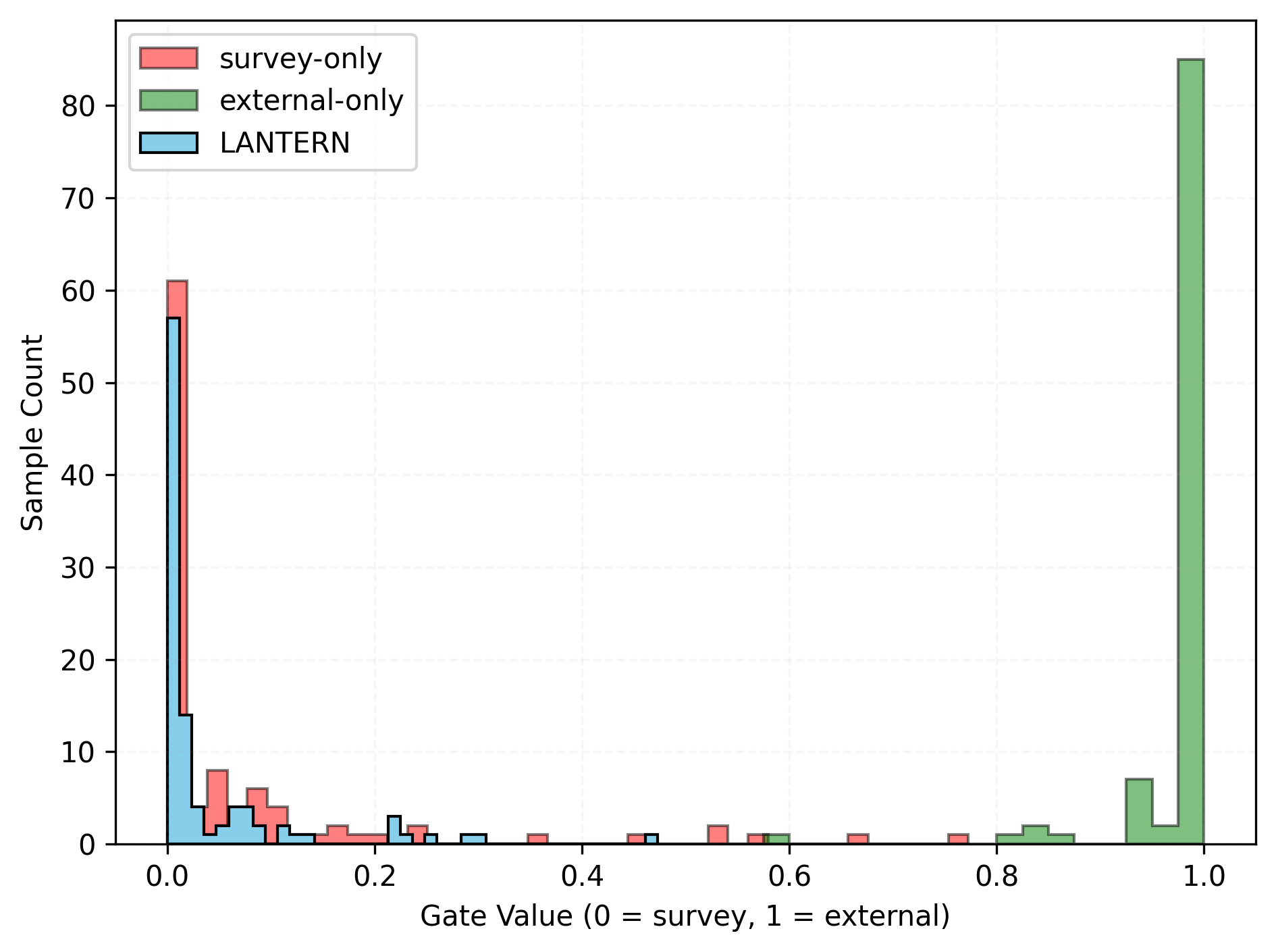}}
\caption{Gating Behavior for 100 user samples: LANTERN learns to commit; indicated with minimal weights in the mid-range. Gate also treats survey-only as a strong anchor with most weights clustering towards 0 which prevents high-volume noisy external signal from dominating predictions.}
\label{gate-behavior}
\end{center}
\vskip -0.2in
\end{figure}

A core component of LANTERN is the gated residual fusion mechanism, which allows the model to decide, per instance and per response key, how much to incorporate external features into the final user embedding. The fusion formula is given by:
\begin{align*}
    h_{fused} = h_s + g \odot (h_t - h_s)
\end{align*}
where $g \in (0,1)$ is a learned gate, $h_s$ is the survey embeddings and $h_t$ are the embeddings obtained after attending to external information within the transformer block. 
The gating values are not static; they are learned dynamically and may differ across users and output dimensions. To understand how the model uses this mechanism in practice, we analyzed the distribution of gate values across the test set. The resulting histogram \ref{gate-behavior} reveals a \emph{bimodal} distribution. Most gate values cluster tightly towards 0, reinforcing that LANTERN treats survey embeddings as an anchor. The minimal weight placed on soft interpolation in the mid-range suggests that the model learns to commit rather than hedge. The sparse weights towards 1 validates our assumption of noisy external signals despite their volume. The gating behavior strengthens the premise that there is more intent signal in the survey embeddings and hence should be used as an anchor. It also reinforces the interpretability claims of LANTERN as it is possible to inspect a prediction and identify whether it was driven by the survey or by supplemental context.

This selective gating also acts as a fail safe mechanism. In cases where external signals are noisy or misaligned, the model can down weight them entirely, defaulting to survey-only inference. Conversely, for cold-start users or sparsely answered surveys, the model can defer to external signals. The learned gating distribution therefore provides operational transparency into LANTERN’s decision making logic.

\section{Temporal Generalization Across Survey Cycles}
The deployment setting for LANTERN involves an adaptive survey platform where questions can be added, removed, or updated over time. This makes the task non-stationary, not only in the sense of changing features, but also in the changing set of labels; each corresponding to a unique response key derived from survey options. These response keys serve as the binary targets in LANTERN’s output layer.

In a supervised learning regime, the label space defines the output structure of the model. When this space evolves, as it does in adaptive surveys, a static model trained on an earlier set of response keys becomes misaligned with the inference task. Output units may correspond to deprecated questions, and newer response keys may be unrecognized by the model. This leads to systemic performance degradation: missing predictions on active attributes and falsely scoring obsolete ones.

For example, if the current survey includes a new binary question with options "Yes" and "No," those response keys will not exist in the output space of a previously trained model. Likewise, keys that no longer appear in the survey will continue to produce predictions, introducing irrelevant outputs. This misalignment cannot be resolved through fine-tuning alone since it involves a structural change to the label set. We adopt a synchronous retraining strategy; with each new cycle of the adaptive survey, the system extracts the current set of response keys and triggers a fresh end-to-end training pass. Due to the modular nature of LANTERN’s architecture, encoder components can be reused when appropriate, but the full fusion pipeline and output head are retrained to align precisely with the current label space. This guarantees that the deployed model remains synchronized with the active survey, avoiding drift and ensuring correctness in both training and inference.

Retraining is facilitated by automatic job scheduling and data pipeline versioning. Because the survey evolves on a predictable cadence, the retraining process can be operationalized as part of the model lifecycle without introducing manual intervention.

\end{document}